%% file: acl_latex.tex
\documentclass[11pt]{article}

\usepackage[preprint]{acl}

\usepackage{times}
\usepackage{latexsym}
\usepackage{makecell}

\usepackage[T1]{fontenc}

\usepackage[utf8]{inputenc}

\usepackage{microtype}

\usepackage{inconsolata}

\usepackage{graphicx}
\usepackage{subcaption}

\usepackage{amsfonts}
\usepackage{amsmath}

\usepackage{multirow}
\usepackage{booktabs}
\usepackage[table]{xcolor}
\usepackage{colortbl}  
\usepackage{arydshln}

\title{Bridging the Discrete-Continuous Gap: Unified Multimodal Generation via Coupled Manifold Discrete Absorbing Diffusion}

\author{
  \textbf{Yuanfeng Xu\textsuperscript{1}},
  \textbf{Yuhao Chen\textsuperscript{1}},
  \textbf{Liang Lin\textsuperscript{1,2,3}},
  \textbf{Guangrun Wang\textsuperscript{1,2,3}}\footnotemark[2]
\\
\\
\textsuperscript{1}Sun Yat-sen University,
  \textsuperscript{2}Guangdong Key Lab of Big Data Analysis \& Processing,
  \textsuperscript{3}X-Era AI Lab
  \\
  \small{
    \textbf{Email:} \href{mailto:xuyf93@mail2.sysu.edu.cn}{xuyf93@mail2.sysu.edu.cn}, \href{mailto:wanggrun@gmail.com}{wanggrun@gmail.com}
  }
}

\begin{document}
\maketitle
\renewcommand{\thefootnote}{\fnsymbol{footnote}} 
\footnotetext[2]{Corresponding author.}

\input{sec/0_abstract}
\input{sec/1_intro}

\input{sec/2_relate}

\input{sec/3_method}

\input{sec/4_exp}

\input{sec/5_conclusion}

\input{sec/6_limitation}

\bibliography{custom}

\end{document}

%% file: sec/0_abstract.tex
\begin{abstract}
The bifurcation of generative modeling into autoregressive approaches for discrete data (text) and diffusion approaches for continuous data (images) hinders the development of truly unified multimodal systems. While Masked Language Models (MLMs) offer efficient bidirectional context, they traditionally lack the generative fidelity of autoregressive models and the semantic continuity of diffusion models. Furthermore, extending masked generation to multimodal settings introduces severe alignment challenges and training instability. In this work, we propose \textbf{CoM-DAD} (\textbf{Co}upled \textbf{M}anifold \textbf{D}iscrete \textbf{A}bsorbing \textbf{D}iffusion), a novel probabilistic framework that reformulates multimodal generation as a hierarchical dual-process. CoM-DAD decouples high-level semantic planning from low-level token synthesis. First, we model the semantic manifold via a continuous latent diffusion process; second, we treat token generation as a discrete absorbing diffusion process, regulated by a \textbf{Variable-Rate Noise Schedule}, conditioned on these evolving semantic priors. Crucially, we introduce a \textbf{Stochastic Mixed-Modal Transport} strategy that aligns disparate modalities without requiring heavy contrastive dual-encoders. Our method demonstrates superior stability over standard masked modeling, establishing a new paradigm for scalable, unified text-image generation.

\end{abstract}

%% file: sec/1_intro.tex
\section{Introduction} \label{sec:intro}

The pursuit of Artificial General Intelligence (AGI) necessitates models capable of reasoning and generating across diverse modalities. However, a fundamental topological disconnect persists in current architectures: language is inherently discrete and symbolic, while visual data is continuous and dense. Consequently, the field has fragmented into two dominant paradigms: Autoregressive (AR) models, which excel at discrete text generation, and Continuous Diffusion Models (CDMs), which dominate high-fidelity image synthesis.

\begin{figure}[t]
  \centering
  \includegraphics[width=0.42\textwidth]{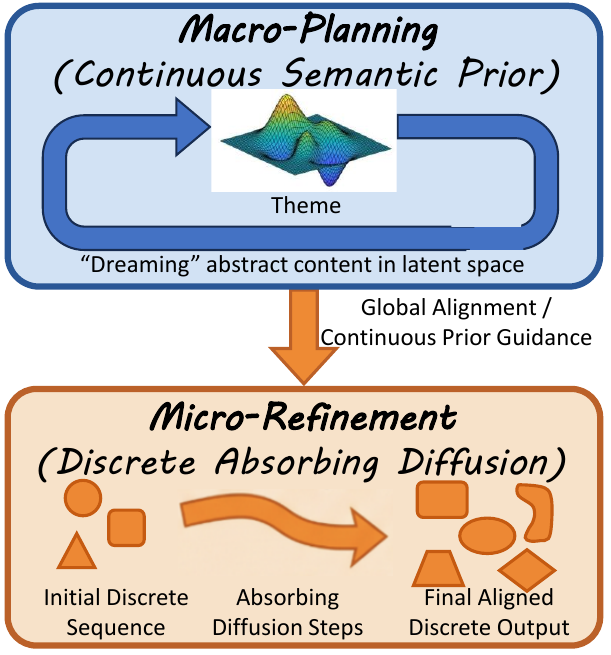}
  \vspace{-6pt}
  \caption{\textbf{Overview of CoM-DAD.} The framework splits generation into \textbf{Macro-Planning} (Top), where a continuous latent diffusion models abstract semantic "themes" (the "dreaming" phase), and \textbf{Micro-Refinement} (Bottom), where a discrete absorbing diffusion synthesizes tokens. The vertical arrow signifies the conditioning of the discrete generation process on the continuous prior, ensuring global alignment between the abstract plan and the final token sequence.}    \label{fig:com_dad_overview}
  \vspace{-24pt}
\end{figure}

Efforts to unify these paradigms often result in compromised hybrids. Masked Generative Models (MGMs) attempt to bridge this gap by treating generation as a parallel denoising task. While MGMs offer significant efficiency gains over AR models (which suffer from serial dependency) and faster inference than CDMs, they notoriously struggle with \textbf{generative consistency}. Without a strong prior, masking-based models often produce locally coherent but globally disjoint outputs. Recent advancements, such as Representation-Conditioned Generation (RCG) \citep{li2024return}, have mitigated this in the visual domain by conditioning pixel generation on self-supervised representations. However, RCG remains strictly unimodal, focusing on unconditional image synthesis and failing to address the complexities of cross-modal alignment or the discrete nature of language.

We argue that the difficulty in training multimodal masked models stems from forcing a single network to simultaneously learn semantic abstraction (what to generate) and structural composition (how to arrange tokens). To resolve this, we introduce CoM-DAD, a hierarchical framework that mathematically formalizes masked generation not as simple “filling in the blanks,” but as a Discrete Absorbing Diffusion Process guided by a Continuous Semantic Prior. As conceptually illustrated in Figure \ref{fig:com_dad_overview}, this hierarchical decoupling allows us to manage semantic abstraction and structural composition on separate, optimized manifolds. Our approach operates on two levels:
\begin{enumerate}
\item Macro-Planning (Latent Space): We employ a lightweight diffusion model to navigate the continuous manifold of high-level semantic representations. This allows the model to “dream” the abstract content of an image or sentence before committing to specific tokens.  \vspace{-6pt}
\item Micro-Refinement (Discrete Space): We formulate token generation as a reverse diffusion process where tokens emerge from an absorbing state ([MASK]). Unlike standard MLMs, our transition kernel is strictly conditioned on the macro-plan, ensuring that every generated token is globally aligned with the intended semantic target. Crucially, this process is governed by a Variable-Rate Noise Schedule, which replaces fixed masking ratios with a continuous time parameter. This allows the model to learn generation from pure noise, while the transition kernel remains strictly conditioned on the macro-plan to ensure global alignment.
\end{enumerate}

Furthermore, to address the scarcity of aligned multimodal data, we propose a \textbf{Stochastic Mixed-Modal Transport} mechanism during training. By dynamically swapping semantic priors between modalities (e.g., forcing the model to generate image tokens from a text representation), we induce a unified semantic space without the need for auxiliary alignment losses like CLIP.

Our contributions can be summarized as follows:\begin{itemize}    \item We propose \textbf{CoM-DAD}, a unified probabilistic framework that bridges topological gaps between modalities by coupling a continuous latent diffusion for semantic planning with a discrete absorbing diffusion for synthesis.    \item We introduce a \textbf{Variable-Rate Discrete Diffusion} mechanism that generalizes masked modeling with a continuous noise schedule, significantly improving generative consistency over static masking strategies.    \item We develop a \textbf{Stochastic Mixed-Modal Transport} strategy that naturally aligns visual and textual manifolds via dynamic representation swapping, eliminating the need for heavy contrastive dual-encoder pre-training.    \item We demonstrate that our hierarchical decoupling achieves superior training stability and sampling efficiency compared to standard autoregressive or monolithic diffusion baselines in multimodal contexts.\end{itemize}

%% file: sec/2_relate.tex
\section{Related Work}
\label{sec:related_works}

\paragraph{Diffusion Models for Discrete Generation.}
Diffusion models have achieved notable success in continuous domains such as image and audio generation~\citep{sohl2015deep, ho2020denoising}. Extending diffusion to discrete sequences has attracted increasing interest~\citep{li2025situ}, particularly for language and symbolic reasoning. Early works~\citep{hoogeboom2021argmax, austin2021structured} adapt diffusion to discrete spaces via relaxation or masking strategies. Later approaches~\citep{li2022diffusion, he2022diffusionbert} embed discrete tokens into continuous spaces to enable Gaussian diffusion. Although effective, this introduces optimization difficulties, as simultaneously optimizing both the embedding layer and the diffusion model can lead to shortcut learning. In contrast, \textbf{CoM-DAD} operates directly on the discrete manifold. By introducing a hierarchical coupling with a continuous latent planner, it enables stable, efficient, and controllable generation without the optimization instability associated with joint embedding-diffusion training.

\begin{figure*}[t]
  \centering
  \includegraphics[width=1.\textwidth]{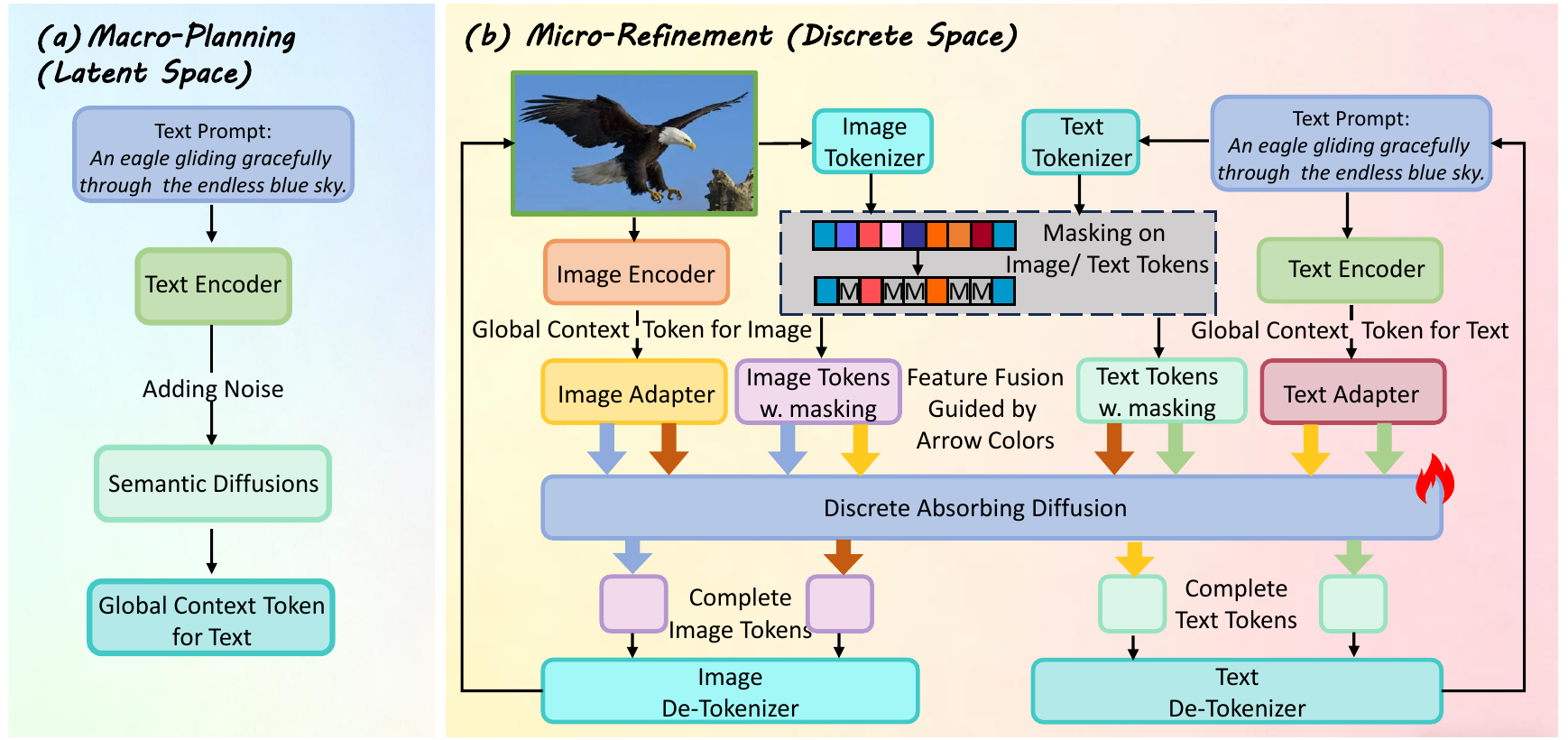}
  \vspace{-24pt}
  \caption{\textbf{The CoM-DAD Training Pipeline.} The framework consists of two coupled diffusion processes. \textbf{Left (Stage I):} The \textit{Manifold-Constrained Semantic Diffusion} learns a continuous prior over semantic representations ($r$) via an SDE, capable of handling both text and image modalities. \textbf{Right (Stage II):} The \textit{Semantic-Aware Discrete Absorbing Diffusion} reconstructs the discrete token sequence ($x$) from a masked state ($\tilde{x}_t$). The \textbf{Semantic Injection Interface} (center) connects these topologies by projecting the sampled semantic plan $r$ into the decoder’s embedding space, conditioning the reverse diffusion step $p_\theta(x_{t-1} | \tilde{x}_t, r)$ to ensure global semantic coherence. Cross-Modal Alignment is applied to (b).}    \label{fig:method_pipeline}
\vspace{-20pt}
\end{figure*}

\paragraph{Masked Language Models as Discrete Absorbing Processes.}Masked language modeling (MLM) has recently been reinterpreted as a form of discrete diffusion with absorbing states~\citep{gemini_diffusion2025, nie2025large, wu2025fast, dream2025}. While these methods effectively approximate autoregressive generation through iterative unmasking, they typically restrict both semantic abstraction and structural composition to the discrete token space. \textbf{CoM-DAD} distinguishes itself by hierarchically decoupling these tasks. We employ a continuous latent diffusion to generate a high-level semantic plan, which serves as a robust condition for the discrete absorbing process, effectively bridging the topological gap between continuous semantic planning and discrete structural generation.

\paragraph{Multimodal Generation and Semantic Guidance.} Multimodal generation aims to produce coherent outputs across heterogeneous data types. Prior work often relies on joint embedding spaces or autoregressive multimodal transformers~\citep{alayrac2022flamingo, chen2023pali, team2023gemini} to bridge the modality gap. \textbf{CoM-DAD} advances this paradigm by introducing Stochastic Mixed-Modal Transport. Rather than treating modalities as disparate sources to be aligned, we unify them into a shared continuous semantic manifold. This allows for dynamic prior swapping, where the discrete diffusion process is universally guided by the continuous planner, facilitating seamless cross-modal generalization. Semantic-Aware generation~\citep{li2024return,wang2022traditional} demonstrates that high-level representations can effectively guide continuous diffusion, though existing methods remain unimodal. \textbf{CoM-DAD} extends this idea to multimodal discrete generation by injecting learned representations directly into the absorbing diffusion process, enabling efficient semantic planning and cross-modal alignment within a unified framework.

%% file: sec/3_method.tex
\section{Method}

In this section, we formally detail \textbf{CoM-DAD}, a hierarchical generative framework that bridges the gap between continuous semantic exploration and discrete token generation. Unlike prior semantic-conditioned approaches such as RCG \citep{li2024return}, which focus exclusively on unimodal image synthesis using standard backbones, CoM-DAD introduces a \textbf{unified discrete diffusion mechanism} capable of joint text-image modeling.

Our framework, schematically illustrated in Figure \ref{fig:method_pipeline}, decomposes the intractable multimodal distribution $p(x)$ into two tractable generative processes operating in distinct topological spaces: (1) a \textbf{Continuous Latent Diffusion} process modeling the high-level semantic manifold $\mathcal{R}$, and (2) a \textbf{Discrete Absorbing Diffusion} process modeling the token-space conditional distribution $p(x|r)$. The figure highlights how the Semantic Injection Interface acts as the critical bridge, translating the continuous "plan" from the latent diffusion into a guiding signal for the discrete token denoiser.

\begin{table*}
  \centering
  \caption{\textbf{Quantitative comparison of unconditional text generation performance.} \textbf{CoM-DAD} outperforms existing autoregressive and masked language model baselines in both BLEU-2 and BLEU-4 metrics. This superior fidelity validates the effectiveness of the Continuous Latent Planner in maintaining global coherence across the discrete text manifold. ``Ours (large) + Autoregressive'' denotes a variant where CoM-DAD is constrained to generate tokens in a sequential order.}\label{tab:main_table}
    \vspace{-12pt}
  \resizebox{1.0\hsize}{!}{
    \begin{tabular}{cccccccc}
        \toprule
        Methods & Pretained  &  \makecell{Training \\Steps} & \makecell{Inference \\Iterations} & \makecell{Output \\Length} & BLEU /\% ($\uparrow$) & Self-BLEU /\% ($\downarrow$) \\
        \midrule
        \multicolumn{7}{c}{\textit{Autoregressive Prediction}} \\\cdashline{1-7}[1pt/1pt]
        GPT-2   & \checkmark      & 1M    & 40    & 40    & 10.81      & 40.02 \\
        BERT (base) \citep{devlin2018bert} & \checkmark       & 1M    & 40    & 40    & 7.80      & 10.06 \\
        BERT (large) \citep{devlin2018bert} & \checkmark       & 1M    & 40    & 40    & 5.05      & \textbf{9.43} \\
         \rowcolor{gray!30} Ours (base) + Autoregressive  & ×           & 400K+300K & 20      & 256      & 8.52      & 18.37 \\
         \rowcolor{gray!30} Ours (large) + Autoregressive  & ×          & 400K+300K & 20      & 256      & \textbf{13.64}      & 16.52 \\
        \midrule
        \multicolumn{7}{c}{\textit{Diffusion-based Methods}} \\\cdashline{1-7}[1pt/1pt]
        D3PM \citep{austin2021structured}  & ×       & 1M    & 128   & 128   & 42.41      & 22.88 \\
        Diffusion-LM \citep{li2022diffusion} & ×        & 740K  & 2000  & 64    & 35.53      & 26.68 \\
        DiffusionBERT \citep{he2022diffusionbert} & ×      & 1.9M  & 128   & 128   & 43.58     & 21.51 \\
        BERT-Mouth \cite{wang2019bert} & \checkmark       & 2.8M  & 10    & 50    & 28.67      & \textbf{12.4} \\
        \rowcolor{gray!30} Ours (base)  & ×          & 400K+300K & 20      & 256      &  29.42     & 18.37 \\
        \rowcolor{gray!30} Ours (large)  & ×          & 400K+300K & 20      & 256      & \textbf{47.46}      & 16.52 \\
        \bottomrule
        \end{tabular}
        }
        \vspace{-18pt}
\end{table*}

\subsection{Theoretical Formulation}
Let $x \in \mathcal{X}$ represent a discrete sequence (e.g., text tokens or quantized image patches) and $r \in \mathbb{R}^d$ be a continuous semantic vector derived from a pre-trained encoder $\mathcal{E}(x)$. We maximize the evidence lower bound (ELBO) of the log-likelihood $\log p_\theta(x)$:
\begin{small}
\begin{equation}
\log p_\theta(x) \geq \underbrace{\mathbb{E}{q(r|x)}[\log p\theta(x|r)]}_{\text{reconstruction}} - \underbrace{D{\text{KL}}(q(r|x) | p_\phi(r))}_{\text{prior matching}}.
\end{equation}
\end{small}

Unlike standard VAEs where the prior $p(r)$ is a static Gaussian, we parameterize $p_\phi(r)$ as a \textbf{continuous diffusion model}. Furthermore, we model the reconstruction term $p_\theta(x|r)$ not as a simple autoregressive decoder, but as a \textbf{discrete diffusion process} over the vocabulary set $V$. This hybrid formulation allows CoM-DAD to decouple global semantic planning (in continuous space $\mathcal{R}$) from local structural refinement (in discrete space $\mathcal{X}$).

\subsection{Stage I: Manifold-Constrained Semantic Diffusion}The first stage models the prior distribution of semantic representations $p_\phi(r)$. While RCG \citep{li2024return} utilizes a standard diffusion model for image class embeddings, we employ a modality-agnostic diffusion process capable of navigating the joint semantic space of both vision and language.

Given a semantic vector $r_0 = \mathcal{E}(x)$, we define a forward stochastic differential equation (SDE) that gradually destroys semantic information:\begin{equation}d r_t = -\frac{1}{2} \beta(t) r_t dt + \sqrt{\beta(t)} d \mathbf{w}_t,\end{equation}where $\mathbf{w}_t$ is standard Brownian motion. We train a time-dependent denoiser $\epsilon_\phi(r_t, t)$ to reverse this process. Crucially, to ensure stability across modalities with varying norms, we employ \textbf{representation normalization} before the diffusion process. The training objective is the standard reweighted variational bound:\vspace{-5pt}\begin{equation}\mathcal{L}_{\text{latent}} = \mathbb{E}_{t, r_0, \epsilon} \left[ \| \epsilon - \epsilon_\phi(r_t, t, c_m) \|^2 \right],\end{equation}
where $c_m$ indicates the modality ID, allowing the single model to learn the distinct manifolds of textual ($r_{txt}$) and visual ($r_{img}$) semantics simultaneously.

\subsection{Stage II: Semantic-Aware Discrete Absorbing Diffusion}\label{sec:DAD}

The core innovation of CoM-DAD is the formulation of sequence generation as a \textbf{Discrete Absorbing Diffusion Process}, generalizing the concept of ``masked language modeling'' into a rigorous probabilistic framework.

\paragraph{Discrete Forward Process.} Let $x_0 = (w_1, \dots, w_L)$ be a sequence of tokens from vocabulary $V$. We define a forward transition matrix $Q_t$ that transitions any token $w$ to a special absorbing state \texttt{[MASK]} with probability $\gamma_t$, and leaves it unchanged with probability $1-\gamma_t$. This defines a corrupted sequence $\tilde{x}_t$ where a subset of tokens are masked according to the Markov property of the absorbing state.\paragraph{Variable-Rate Noise Schedule.} A critical component of our approach is the noise schedule $\gamma(t)$. Unlike the fixed masking strategies used in standard BERT-like models (typically 15\%), we sample the masking ratio $\gamma_t$ from a continuous time schedule $t \sim U[0,1]$. This creates a \textbf{Variable-Rate Noise Schedule} that forces the model to learn generation across the entire complexity spectrum—from pure noise ($\gamma_1 \approx 1$) to fine-grained refinement ($\gamma_0 \approx 0$). This dynamic schedule is what effectively shifts the model's capability from simple local infilling to robust global generation.

\paragraph{Semantic-Aware Denoising.} We learn a reverse transition kernel $p_\theta(x_{t-1} | \tilde{x}_t, r)$ parameterized by a Transformer. To condition the discrete generation on the continuous semantic vector $r$ sampled from Stage I, we introduce a \textbf{Semantic Injection Interface}:
\vspace{-12pt}
\begin{equation}h_0 = [\text{Proj}(r); \text{Embed}(\tilde{x}_t)].\end{equation}
\vspace{-3pt}
By projecting $r$ into the token embedding space and prepending it as a global context token, the Transformer attention mechanism allows every discrete decoding step to attend to the global semantic plan. The learning objective is the negative log-likelihood over the masked regions $\mathcal{M}$:\begin{equation}\mathcal{L}_{\text{discrete}} = \mathbb{E}_{x, t, r} \left[ - \sum_{i \in \mathcal{M}} \log p_\theta(x_i | \tilde{x}_{\setminus \mathcal{M}}, r) \right].\end{equation}

This formulation fundamentally differs from RCG, which relies on pixel-space diffusion or latent VAE decoders. By operating in discrete token space, it natively handles text and quantized images (via VQ-VAE tokens) in a unified architecture.\

\subsection{Cross-Modal Alignment via Inter-Modal Optimal Transport}\label{sec:aligment}

A critical challenge in multimodal generation is the misalignment between visual and textual representation spaces. We address this via a \textbf{Mixed-Modal Sampling Strategy} that effectively approximates an optimal transport plan between modalities.

We construct training batches $\mathcal{B} = \{ \mathcal{B}_{txt}, \mathcal{B}_{img}, \mathcal{B}_{pair} \}$.\begin{itemize}    \item \textbf{Intra-Modal Learning:} For $\mathcal{B}_{txt}$ and $\mathcal{B}_{img}$, we train the model to reconstruct $x$ given its own representation $r = \mathcal{E}(x)$.    \item \textbf{Cross-Modal Bridge:} For paired data $\mathcal{B}_{pair}$, we perform \textbf{representation swapping}. We train the model to generate image tokens $x_{img}$ conditioned on text representations $r_{txt}$, and vice-versa.\end{itemize}To facilitate this, we introduce lightweight \textbf{Modality Adapters} $\mathcal{A}_{T \to V}$ and $\mathcal{A}_{V \to T}$ that project representations into a shared semantic centroid before injection. This forces the latent diffusion model (Stage I) and the discrete generator (Stage II) to agree on a unified semantic coordinate system, enabling zero-shot generation (e.g., text-to-image) even with limited paired data.

%% file: sec/4_exp.tex
\section{Experiments}
\label{sec:experiments}

\begin{figure*}[t]
  \centering
  \begin{subfigure}[t]{0.45\textwidth}
    \centering
    \includegraphics[width=\linewidth]{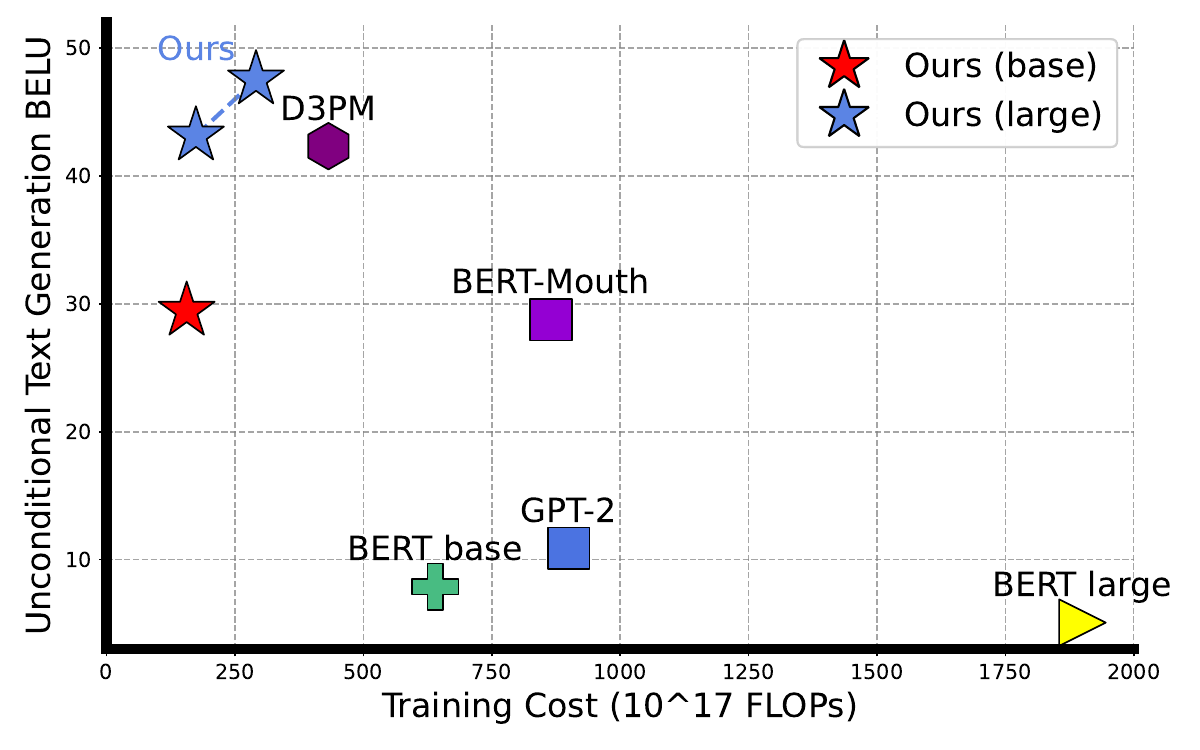}
\vspace{-18pt}
    \caption{Training FLOPs versus BLEU performance.}
    \label{fig:scatter}
  \end{subfigure}
\vspace{-5pt}
  \hfill
  \begin{subfigure}[t]{0.45\textwidth}
    \centering
    \includegraphics[width=\linewidth]{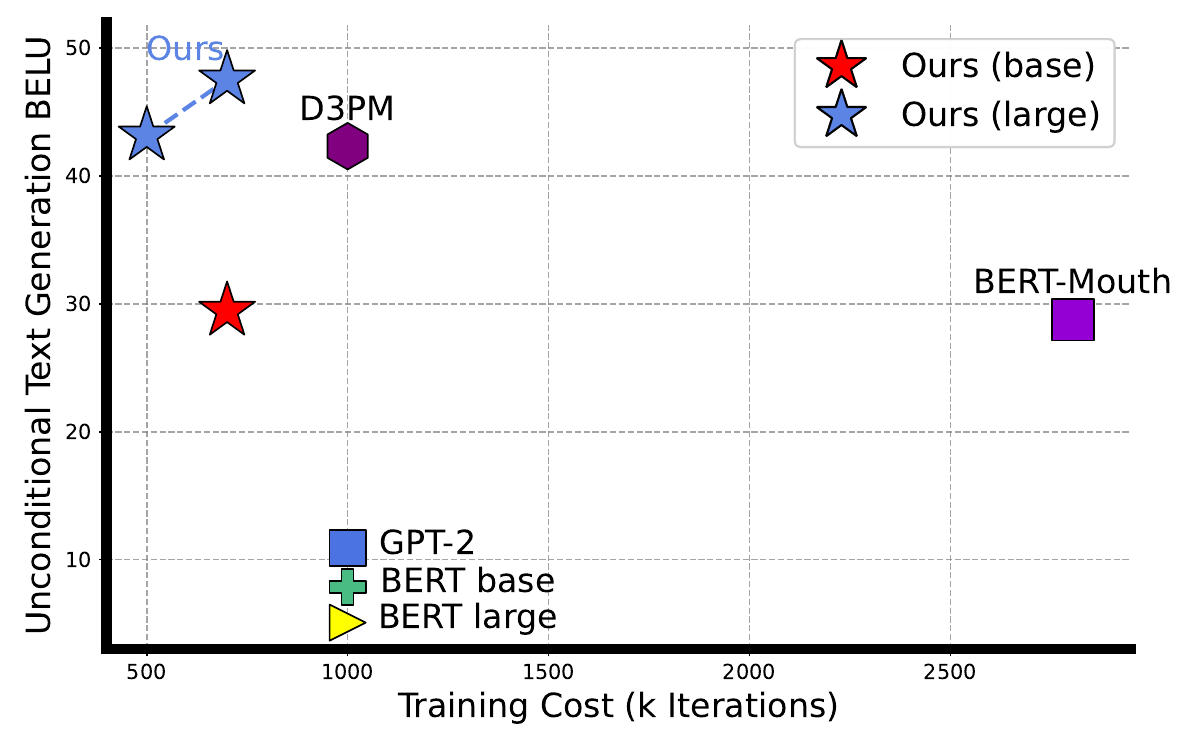}
\vspace{-18pt}
    \caption{Convergence behavior across training iterations.}
    \label{fig:scatter_iter}
\vspace{-5pt}
  \end{subfigure}
  \caption{\textbf{Impact of the Semantic Injection Interface on convergence efficiency.} \textbf{CoM-DAD} achieves superior BLEU scores with substantially reduced training costs compared to the ablated variant without the interface. The reported cost includes training of both the \textit{Continuous Latent Planner} (Stage~I) and \textit{Discrete Absorbing Diffusion} (Stage~II). These results indicate that the \textbf{Semantic Injection Interface} accelerates convergence by enabling the discrete model to exploit the structure of the continuous semantic manifold, rather than learning it from scratch.}
  \label{fig:semantic_injection}
\vspace{-18pt}
\end{figure*}

We empirically evaluate the effectiveness of \textbf{CoM-DAD} on both unimodal and cross-modal generative tasks. Our goals are threefold: (1) to assess the generative quality and efficiency of the discrete absorbing process on high-dimensional image and text manifolds, (2) to validate the hierarchical decoupling hypothesis by analyzing the impact of the Continuous Latent Planner and Semantic Injection Interface on convergence and stability, and (3) to investigate the efficacy of Stochastic Mixed-Modal Transport for zero-shot cross-modal alignment and generalization.

\subsection{Implementation Details}\label{implementation_details}

\paragraph{Data Sources and Mixed-Modal Transport.} Following our \textbf{Stochastic Mixed-Modal Transport} strategy (Sec.~\ref{sec:aligment}), we construct a unified training distribution comprising three subsets: (i) large-scale textual data from BookCorpus and Wikipedia~\citep{wettig2022should} (28M samples), (ii) image-only data from ImageNet-1k (1.28M samples), and (iii) 100K image-text pairs curated from COCO, with synthetic captions generated using Chameleon~\citep{team2024chameleon}. To ensure balanced manifold coverage and stable prior swapping, the sampling ratio across these sources is fixed at 2:2:1.
\vspace{-12pt}

\paragraph{Discrete Manifold Tokenization.} To establish the discrete state space, images are resized and center-cropped to $256 \times 256$ pixels, then tokenized using VQGAN~\citep{yu2021vector} into 256 discrete visual tokens. Text inputs are tokenized using the RoBERTa tokenizer~\citep{liu2019roberta} and padded or truncated to a maximum of 256 tokens. No additional data augmentation is applied, relying on the variable-rate noise schedule for robustness.
\vspace{-12pt}
\paragraph{Continuous Manifold and Optimization.}We utilize frozen self-supervised encoders to define the continuous semantic manifold: MoCoV3~\citep{chen2021empirical} for images and MPNet for text. The framework is trained in two decoupled stages:\begin{itemize}    
\item \textbf{Stage I (Continuous Latent Diffusion):} The continuous planner is trained 400K iterations to model the density of semantic embeddings.    \vspace{-6pt}\item \textbf{Stage II (Discrete Absorbing Diffusion):} The discrete generator is trained 300K iterations to generate tokens from absorbing states.\end{itemize}We use the AdamW optimizer with an initial learning rate of $5 \times 10^{-4}$. All experiments are conducted on 8 NVIDIA A800 GPUs.
\vspace{-10pt}

\paragraph{Evaluation Protocol.}We evaluate the topological unification capabilities of CoM-DAD. For unconditional image generation, we follow~\citep{li2024return} and generate 50K samples from the ImageNet distribution, reporting Inception Score (IS) \citep{salimans2016improved} and Fréchet Inception Distance (FID) \citep{heusel2017gans}. For text generation, we report BLEU (n-gram accuracy) \citep{papineni2002bleu} and Self-BLEU (diversity) \citep{zhu2018texygen}. For cross-modal generation, we test the Modality Adapters by prompting with text to synthesize aligned images, evaluating semantic consistency through qualitative visual inspection.

\subsection{Main Results and Analysis}\label{main_results}

\paragraph{Superior fidelity on discrete text manifolds.}
We first evaluate \textbf{CoM-DAD} on unconditional text generation and compare it with strong baselines, including autoregressive models and standard masked language models. Table~\ref{tab:main_table} shows that CoM-DAD achieves BLEU-2 and BLEU-4 scores of 47.46 and 13.64, respectively, outperforming all prior approaches under comparable settings. We also include a variant labeled ``Ours (large) + Autoregressive'', where CoM-DAD operates in a sequential manner, demonstrating the framework's flexibility to encompass autoregressive generation as a special case. Compared to standard autoregressive models such as GPT-2, CoM-DAD demonstrates exceptional stability in generating long-range dependencies from scratch, validating the efficacy of decoupling semantic planning from token generation. The generated text exhibits not only local fluency but also superior global coherence, a direct result of the Continuous Latent Planner governing the generation trajectory.

\paragraph{Semantic Injection Interface facilitates convergence.} Figure~\ref{fig:semantic_injection} analyzes the relationship between generation quality and training cost. CoM-DAD achieves superior convergence rates with significantly fewer iterations than an ablated variant lacking the \textbf{Semantic Injection Interface}. This indicates that conditioning the discrete absorbing process on the continuous semantic manifold allows the model to bypass the difficulty of learning structure from scratch. Furthermore, models utilizing this interface produce longer and semantically richer sequences, whereas unguided counterparts tend to suffer from mode collapse or repetition, failing to bridge the gap between the discrete and continuous states effectively.

\begin{figure*}[t]
  \centering
  \begin{minipage}[t]{0.55\textwidth}
    \centering
    \includegraphics[width=\linewidth]{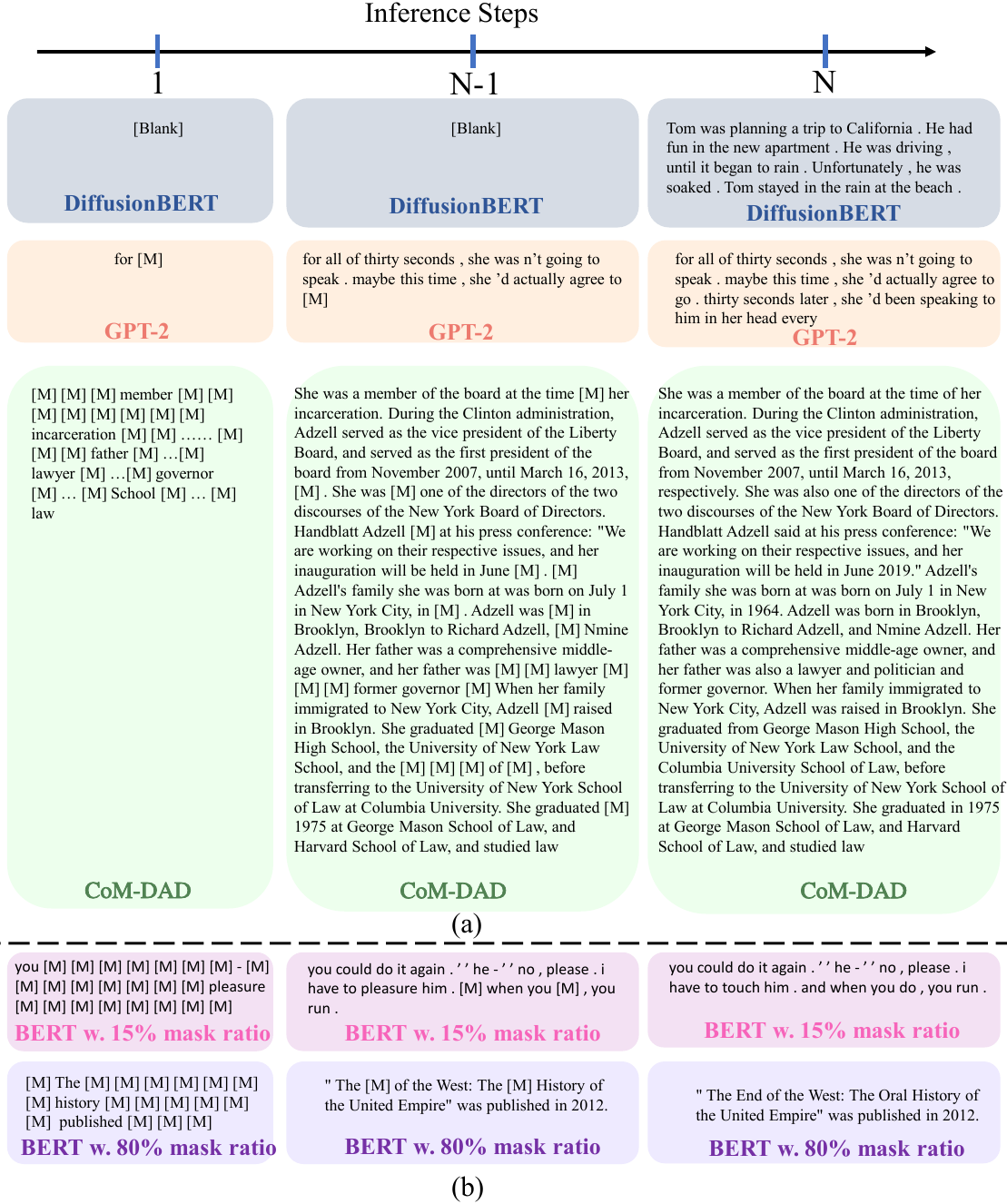}
    \vspace{-18pt}
    \caption{\textbf{Analysis of parallel decoding dynamics and schedule impact.} 
    (a) Comparison of generation paradigms: \textbf{CoM-DAD} utilizes 
    \textit{Discrete Absorbing Diffusion} for efficient non-autoregressive 
    parallel decoding, contrasting with the serial nature of autoregressive 
    baselines. (b) Ablation on absorbing rates: Models trained with the aggressive 
    \textbf{Variable-Rate Noise Schedule} (High Masking) demonstrate emergent 
    semantic prioritization (\emph{main-first, details-later}), establishing global 
    structure before local details. Zoomed-in views are provided for clarity.}
    \label{fig:demo_text_1}
  \end{minipage}
  \hfill
  \begin{minipage}[t]{0.35\textwidth}
    \centering
    \includegraphics[width=\linewidth]{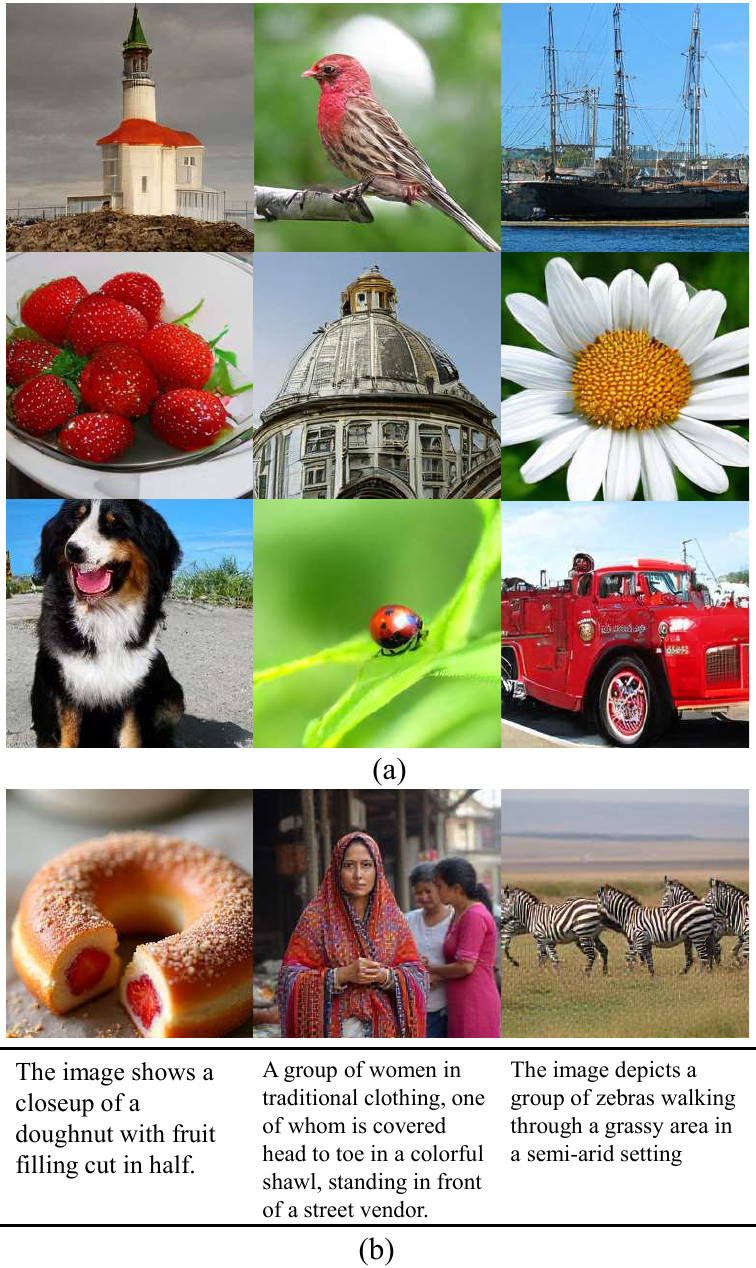}
    \vspace{-18pt}
    \caption{\textbf{Visualization of D-MLLM’s image generation capabilities.} 
    (a) Unconditional image generation results demonstrating diversity and visual 
    quality. (b) Text-to-image generation showcasing effective cross-modal 
    alignment, with synthesized images accurately reflecting the semantic content 
    of input text prompts.}
    \label{fig:demo_image}
  \end{minipage}
  \vspace{-18pt}
\end{figure*}

  \vspace{-10pt}
\paragraph{Parallel decoding via Discrete Absorbing Diffusion.} Unlike autoregressive models that are bound by serial token decoding, CoM-DAD leverages a Discrete Absorbing Diffusion process that allows for non-autoregressive parallel generation. As shown in Figure~\ref{fig:demo_text_1}(a), the model can reconstruct up to 20 tokens per step simultaneously, significantly reducing inference latency. Despite the absence of aggressive GPU-specific optimizations found in mature autoregressive systems, CoM-DAD achieves a $5\times$ speedup over GPT-2. This efficiency confirms that modeling generation as an iterative denoising process from absorbing states is a viable and high-throughput alternative to standard causal modeling.

 \vspace{-10pt}
\paragraph{Emergent semantic prioritization via Variable-Rate Noise Schedule.}An interesting emergent behavior observed in CoM-DAD is its preference for resolving high-information content in early diffusion steps, followed by lower-entropy details—a pattern we term \emph{main-first, details-later}. In Figure~\ref{fig:demo_text_1}(a), the Variable-Rate Discrete Noise Schedule forces the model to first anchor key subject-verb structures before filling in modifiers and syntactic connectors. This validates our hypothesis that the noise schedule dictates the hierarchy of generation.

To isolate this effect, we compare models trained with different absorbing rates (Figure~\ref{fig:demo_text_1}(b)). Only models trained with an aggressive noise schedule inherent to CoM-DAD develop this prioritized behavior, producing more globally consistent outputs. This confirms that the variable-rate schedule does not merely add noise, but actively encourages the model to learn a hierarchical data decomposition.

\begin{table}[t]
  \centering
  \caption{\textbf{Quantitative comparison of unconditional image generation.} \textbf{CoM-DAD} achieves competitive performance against state-of-the-art baselines on the continuous image manifold. These results validate the framework's \textit{Topological Unification}, demonstrating that the \textit{Stochastic Mixed-Modal Transport} strategy effectively aligns discrete token generation with continuous visual semantics.}
  \vspace{-6pt}
\label{tab:quan_image_generation}
  \resizebox{.5\textwidth}{!}{
    \begin{tabular}{cccc}
    \toprule
    Unconditional Generation & params & FID ($\downarrow$) & IS ($\uparrow$) \\
    \midrule
    BigGAN \citep{brock2018large} & 70M   & 38.61 & 24.7 \\
    IC-GAN \citep{casanova2021instance} & 75M   & 15.6  & 59 \\
    ADM \citep{dhariwal2021diffusion}   & 554M  & 26.21 & 39.7 \\
    ADDP \citep{tian2023addp}  & 176M  & 8.9   & 95.3 \\
    MaskGIT \citep{chang2022maskgit} & 227M  & 20.72 & 42.1 \\
    RDM-IN \citep{blattmann2022retrieval} & 400M  & 5.91  & 158.8 \\
    MAGE-B \citep{li2023mage} & 176M  & 8.67  & 94.8 \\
    MAGE-L \citep{li2023mage} & 439M  & 7.04  & 123.5 \\
    RCG-B \citep{li2024return} & 239M  & 3.98  & 177.8 \\
    RCG-L \citep{li2024return} & 502M  & 3.44  & 186.9 \\
    \rowcolor{gray!30} Ours (base) & 318M  & 5.14      & 138.3 \\
    \rowcolor{gray!30} Ours (large) & 593M  & 4.32      & 151.6 \\
    \bottomrule
    \end{tabular}
    }
    \vspace{-18pt}
\end{table}

\begin{figure}[t]
  \centering
  \begin{subfigure}[t]{0.5\textwidth}
    \centering
    \includegraphics[width=\linewidth]{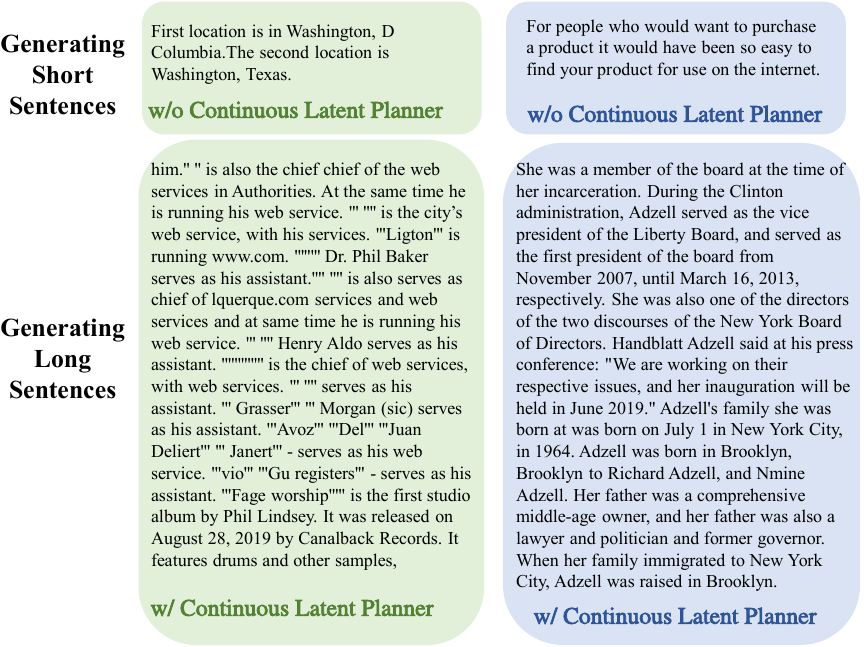}
    \caption{Ablation analysis of CoM-DAD components.}
    \label{fig:ablation_1}
  \end{subfigure}
  \hfill
  \begin{subfigure}[t]{0.5\textwidth}
    \centering
    \includegraphics[width=\linewidth]{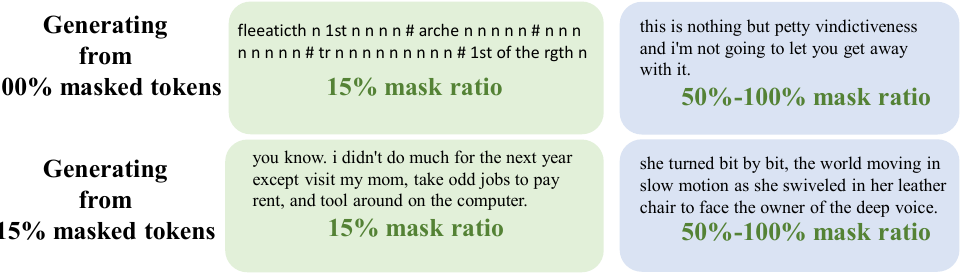}
    \caption{Impact of Noise Schedules.}
    \label{fig:ablation_2}
  \end{subfigure}
    \vspace{-12pt}
  \caption{\textbf{Ablation studies of CoM-DAD.} 
  (a) Removing the \textbf{Continuous Latent Planner} (and the corresponding 
  \textbf{Semantic Injection Interface}) substantially degrades the model’s 
  ability to generate complex, long-form passages, leading to slower convergence 
  and reduced global coherence. This supports our hypothesis that topological 
  decoupling simplifies the learning objective for the 
  \textbf{Discrete Absorbing Diffusion} process. 
  (b) Comparison of the \textbf{Variable-Rate Noise Schedule} against a fixed 15\% rate. The fixed-rate variant yields repetitive, trivial outputs when starting from a \textit{fully absorbing state}. In contrast, CoM-DAD’s variable schedule enables generation from scratch, proving that dynamic masking is critical to shift from local infilling to global generation.} 
  \label{fig:ablation}
  \vspace{-15pt}
\end{figure}
  \vspace{-5pt}
\paragraph{Topological unification and Mixed-Modal Transport.} We evaluate CoM-DAD on unconditional and text-driven image generation to assess its cross-modal capabilities. As shown in Table~\ref{tab:quan_image_generation}, CoM-DAD achieves competitive Inception Scores (IS) and Fréchet Inception Distances (FID), specifically demonstrating that the Stochastic Mixed-Modal Transport strategy successfully aligns heterogeneous manifolds without degradation. 

Figure~\ref{fig:demo_image} visualizes outputs from CoM-DAD. Unconditional samples (Figure~\ref{fig:demo_image}(a)) are visually coherent, while text-driven samples (Figure~\ref{fig:demo_image}(b)) exhibit strong semantic alignment with the prompts. These results demonstrate that our unified continuous semantic manifold allows for effective Dynamic Prior Swapping, enabling the discrete image generator to be accurately guided by textual plans without requiring massive paired datasets.

\vspace{-5pt}
\subsection{Ablation Studies}

To further understand the role of each component in the \textbf{CoM-DAD} framework, we conduct a series of ablations focusing on the hierarchical decoupling and the diffusion noise schedules.
\vspace{-12pt}

\paragraph{Impact of the Continuous Latent Planner.} To validate the necessity of topological decoupling, we remove the \textit{Continuous Latent Planner} (Stage I) and train the \textit{Discrete Absorbing Diffusion} model directly on token sequences without the \textit{Semantic Injection Interface}. As illustrated in Figure~\ref{fig:ablation_1}, models trained without continuous latent conditioning can produce short, syntactically correct phrases but often fail when tasked with generating longer or semantically complex passages. Furthermore, they require significantly more training iterations to achieve comparable quality. This supports our fundamental hypothesis that externalizing semantic planning into a continuous manifold simplifies the discrete learning objective, allowing the token generator to focus solely on mapping high-level plans to discrete structures.
\vspace{-12pt}

\paragraph{Necessity of the Variable-Rate Noise Schedule.} We retrain CoM-DAD using a standard fixed 15\% masking rate (typical of BERT-style MLMs). As shown in Figure~\ref{fig:ablation_2}, this variant fails to generate coherent text from the fully absorbing state, instead repeating simple or semantically trivial tokens. In contrast, the \textbf{Variable-Rate Noise Schedule} employed in CoM-DAD successfully generates complete sentences from scratch. This confirms that aggressive, variable-rate masking is essential for shifting the model from a local infilling objective to a global generative task. These findings are consistent with our insight in Sec.~\ref{sec:DAD} and provide empirical support for \textit{Discrete Absorbing Diffusion} as a mechanism for robust generation rather than mere masked prediction.

%% file: sec/5_conclusion.tex
\section{Conclusion} \label{sec:conclusion}

\paragraph{Summary of Benefits.} CoM-DAD's architectural and training design confers several benefits: (1) \textbf{Efficiency:} By decoupling representation modeling from token generation, CoM-DAD enables faster convergence and $5\times$ faster inference over standard denoising or autoregressive models. (2) \textbf{Generative Capability:} The variable masking schedule fosters the model’s ability to generate coherent and diverse outputs rather than merely recover corrupted inputs. (3)\textbf{Multimodal Alignment:} Our mixed sampling strategy facilitates scalable training from unimodal data while achieving strong cross-modal consistency. (4)\textbf{Unified Architecture:} A single encoder-decoder model handles both text and image generation through a shared conditioning mechanism, supporting flexible and generalizable generation tasks.

%% file: sec/6_limitation.tex
\clearpage
\section*{Limitations} \label{sec:limitation}
While CoM-DAD effectively bridges the topological gap between discrete and continuous modalities, our current empirical validation is primarily focused on foundational image-text generation tasks. Although the Stochastic Mixed-Modal Transport framework is theoretically extensible to temporal modalities like video or audio, we reserve the specific calibration of the Variable-Rate Noise Schedule for these high-dimensional domains for future work to maintain focused analysis. Furthermore, we observe that standard automated metrics may not fully capture the long-horizon semantic consistency driven by the Continuous Latent Planner, potentially underrepresenting the model’s ability to generate conceptually accurate but structurally diverse outputs compared to rigid token-matching baselines.